\documentclass[10pt,twocolumn,letterpaper]{article}

\usepackage{iccv}
\usepackage{times}
\usepackage{epsfig}
\usepackage{graphicx}
\usepackage{amsmath}
\usepackage{amssymb}
\usepackage{capt-of}

\usepackage{graphicx}
\usepackage{subfigure,multirow}
\usepackage{booktabs,tabularx} % for professional tables
 \usepackage{url}            % simple URL typesetting
 \usepackage{amsfonts}       % blackboard math symbols
 \usepackage{microtype}      % microtypography
 \usepackage{color}
 \usepackage{amsmath}
\usepackage{rotating}
\usepackage{stfloats}
\usepackage[normalem]{ulem}
\usepackage{array}
\newcolumntype{M}[1]{>{\centering\arraybackslash}m{#1}}
\newcolumntype{P}[1]{>{\centering\arraybackslash}p{#1}}

\newcommand{\veryshortarrow}[1][3pt]{\mathrel{%
   \vcenter{\hbox{\rule[-.5\fontdimen8\textfont3]{#1}{\fontdimen8\textfont3}}}%
   \mkern-4mu\hbox{\usefont{U}{lasy}{m}{n}\symbol{41}}}}

\usepackage[pagebackref=true,breaklinks=true,letterpaper=true,colorlinks,bookmarks=false]{hyperref}

\iccvfinalcopy 

\def\iccvPaperID{2633}

\ificcvfinal\fi
\begin{document}

\title{Bidirectional One-Shot Unsupervised Domain Mapping}

\author{Tomer Cohen\\
Tel Aviv University\\
{\tt\small tomer104@gmail.com}
\and
Lior Wolf\\
Facebook AI Research and Tel Aviv University\\
{\tt\small wolf@cs.tau.ac.il}
}

\maketitle

%%%%%%%%% ABSTRACT
\begin{abstract}
    We study the problem of mapping between a domain $A$, in which there is a single training sample and a domain $B$, for which we have a richer training set. The method we present is able to perform this mapping in both directions. For example, we can transfer all MNIST images to the visual domain captured by a single SVHN image and transform the SVHN image to the domain of the MNIST images. Our method is based on employing one encoder and one decoder for each domain, without utilizing weight sharing. The autoencoder of the single sample domain is trained to match both this sample and the latent space of domain $B$. Our results demonstrate convincing mapping between domains, where either the source or the target domain are defined by a single sample, far surpassing existing solutions. Our code is made publicly available at \url{https://github.com/tomercohen11/BiOST}.
\end{abstract}

%%%%%%%%% BODY TEXT
\section{Introduction}

There are multiple well-known gaps between many of the current learning techniques and the requirements of ecological cognition, i.e., the ability to operate freely in the real world. These include: (i) ecological learning systems are required to rely mostly on very weakly supervised or unsupervised training data; (ii) the amount of data available during training is small; and (iii) ecological systems should be able to learn continuously, improving over time. 

In this work, we address unsupervised learning in the one-shot scenario, in a way that links the one-sample $x$ to a second, larger training set $S$. The training set $S$ can represent, for example, an existing body of accumulated knowledge. The unsupervised task that we study is cross domain mapping between two visual domains $A$ and $B$, i.e., the ability to translate an image in one visual domain to another visual domain. We present a solution that can perform this mapping in both directions: we can map the single image $x\in A$ to the domain $B$ defined by $S$, and we can also translate any image in domain $B$ to the domain $A$. 

In this scenario, our knowledge of domain $A$ is limited to the single example $x$. This has a few implications, especially when mapping into this domain. At the principled level, we are heavily influenced by this one sample, and our view of domain $A$ is myopic. At the technical level, we are challenged by the inability to train a GAN for domain $A$.  

Our method trains an autoencoder for each domain. The autoencoeder of domain $B$ is first pretrained and then jointly trained with an autoencoder for domain $A$. Despite the limited number of samples in the low-shot domain, weight sharing is not used. Instead, we rely on a feature cycle loss that ensures alignment between the domains, while allowing more freedom to the autoencoders. In addition, we do not employ GANs, not even for domain $B$.

In the literature, one-shot translation was only demonstrated successfully for the mapping from $A$ to $B$ (see Sec.~\ref{sec:prev}). Our success in mapping from $B$ to $A$ provides new insights on unsupervised cross-domain translation. It demonstrates that the variability in the target visual domain does not need to be estimated and it strengthens the link between cross-domain visual translation and style transfer, since style transfer is often applied based on a single image. 

Note, however, that style transfer solutions: (i) typically employ pretrained networks to obtain a perceptual similarity score; (ii) do not benefit from having more than one image from the domain of the content image; and (iii) employ a large style image, while our method can work with low-res visual domains, as well as with high-res ones. The success of our method also demonstrates the emergence of disentanglement between style and content, since the content is translated, while the style is taken from the image $x$. However, unlike the existing methods in the literature, a single encoding pathway is used, i.e., we do not have a style encoder that is separate from the content one.

The new method greatly outperforms the existing algorithms in the one-shot case, in both directions. In comparison to both domain transfer methods and style transfer methods, our method is able to better maintain content. In addition, when mapping to the one-shot domain, the style is faithfully extracted from this one sample, while creating less distortions than previous work.

\section{Related work}
\label{sec:prev}

\begin{figure*}[t]
\centering
\begin{tabular}{c@{~~~~~~~~}c@{~~~~~~~~}c}
   \includegraphics[clip,width=0.289314895\linewidth]{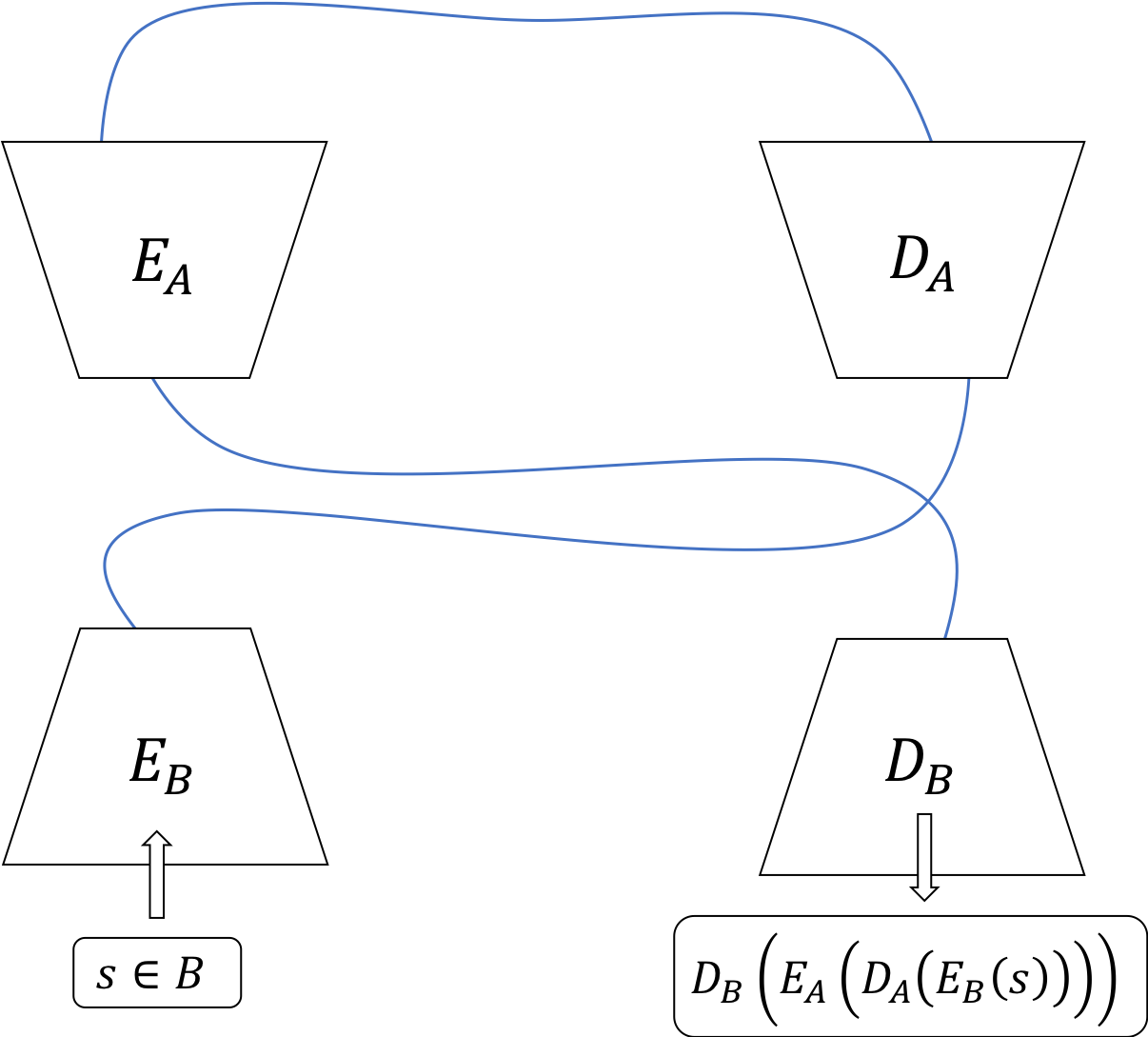}  & 
   \includegraphics[clip,width=0.289314895\linewidth]{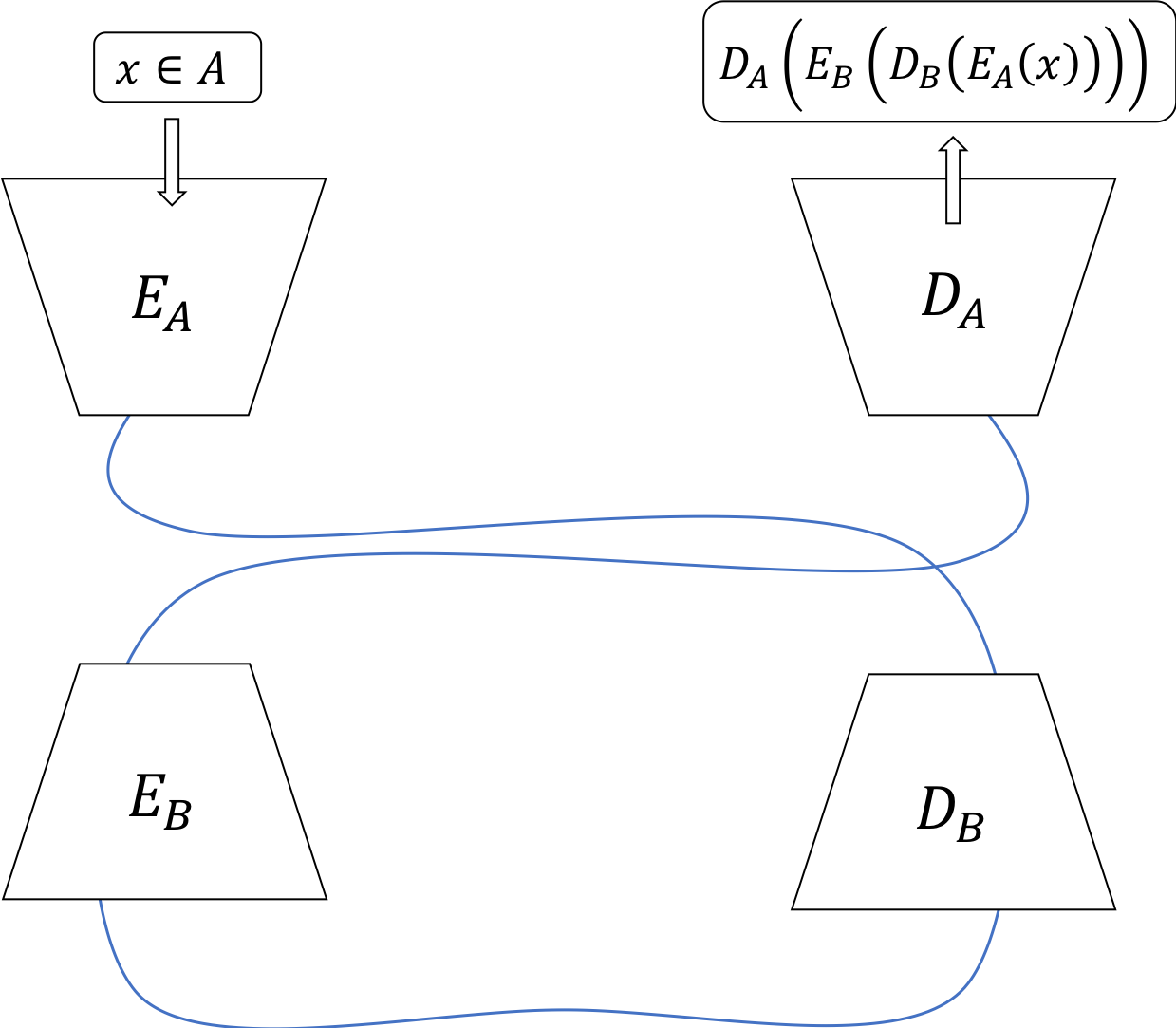}  & 
   \includegraphics[clip,width=0.289314895\linewidth]{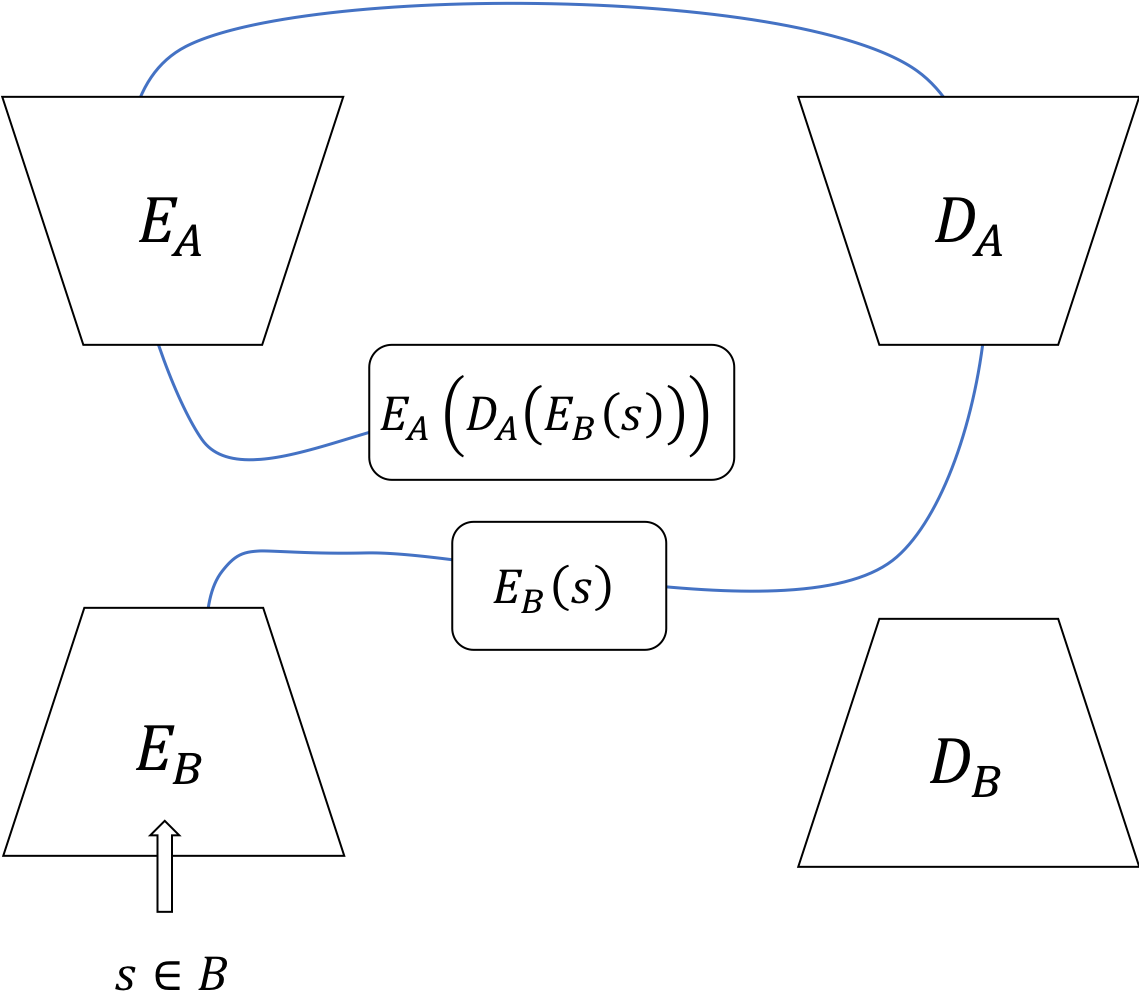}  \\
   (a) & (b) & (c)\\
   \end{tabular}
  \caption{\label{fig:arch} An illustration of the three cycle loss terms used in our method. (a) $\mathcal{L}_{\text{bab-cycle}}$ (b) $\mathcal{L}_{\text{aba-cycle}}$ (c) $\mathcal{L}_{\text{f-cycle}}$. }
\end{figure*}

The field of unsupervised learning has shifted much of its focus from the classical unsupervised tasks of clustering and density estimation to tasks that involve the generation of new samples. This was mainly through the advent of GANs~\cite{gan}, which also led to the development of alternative methods~\cite{glo} and to a renewed interest in autoencoders~\cite{Rumelhart1986LearningRB}. 

The specific task we study is cross domain translation, which is the task of generating an image in the target domain that is conditioned on an image from a source domain. While the supervised case, where the training set consists of matching samples of input/output images, is of considerable practical interest~\cite{pix2pix,pix2pixhd}, in many cases, such samples are very challenging to collect. Unsupervised domain translation methods receive a training set of unmatched samples, one set of samples from each domain, and learn to map between a sample in one domain and the analogous sample in the other domain~\cite{CycleGAN2017,discogan,dualgan,distgan,cogan,unit,stargan,conneau2017word,zhang2017adversarial,zhang2017earth,lample2018unsupervised,munit,i2i_cvpr_2018}. 

The first unsupervised methods learned a mapping in both directions (from $A$ to $B$ and back), in order to apply the circularity constraint: mapping a sample from $A$ to $B$ and then back from $B$ to $A$, should result in the identity function. However, this constraint is unnecessary, and some contributions~\cite{distgan,nam} learn to map in a single direction in an asymmetric way. 

Highly related to this work, is the one-shot translation  method (OST) recently proposed~\cite{ost}. In this work, a single sample $x\in A$ is mapped to the visual domain $B$. However, they are unable to perform the mapping in the other direction {as successfully}, and they write ``the added difficulty in this other direction is that adversarial training cannot be employed directly on the target domain, since only one sample of it is seen''. As we demonstrate, we are able to perform this translation successfully and also greatly outperform OST in the direction they report results on.

Both our method and OST employ one autoencoder per domain, and assume that the latent encoding in the two domains are similar. This basic methodology, which was first introduced by CoGAN~\cite{cogan} is also employed by methods such as UNIT~\cite{unit} and MUNIT~\cite{munit}. These methods~\cite{ost,unit,munit,cogan} all assume a specific type of weight sharing, in which the layers near the image domains (the bottom of the encoders and the top of the decoders) are unshared, while the layers near the latent space are shared. In other words, the top layers of the encoder are shared between the encoders of the two domains and so are the bottom layers of the decoders. In this way, the latent representations of each sample are obtained and processed in a similar way, regardless of the domain of the sample. 
In our work, we do not rely on weight sharing, which we found to be detrimental for the one-shot scenario. Instead, we use a feature-cycle consistency term to ensure the one-shot domain correctly aligns to the multi-shot domain, while the privilege of unshared weights allows the encoders and decoders to learn better transformations, see Sec.~\ref{sec:fcycle}.   

As mentioned, our method further blurs the line between cross domain translation methods and style transfer methods. Style transfer methods~\cite{styletransfer,ulyanov16texture,Johnson2016Perceptual}, synthesize a new image by minimizing the content loss with respect to the input sample, and the style loss with respect to one or more style image samples. The content loss is typically a perceptual loss, i.e., it uses the encoding of the input and the output images by a network pretrained for an image classification task. The style loss compares the statistics of the activations in various layers of this neural network between the output image and the style images. This dual goal was initially obtained by employing a slow optimization process~\cite{styletransfer}. The process was later replaced by feed-forward methods that are trained to produce images that miminize the loss~\cite{ulyanov16texture,Johnson2016Perceptual}. Note that our method, adhering to the unsupervised protocol, does not employ a pretrained classification network.

When translating an image from domain $B$ to the domain defined by $x$, using our method, we would like the image to resemble $x$ in style, while preserving the content of the input image. However, style transfer methods are targeted toward creating visually pleasing images with a certain texture and are dramatically outperformed by cross domain translation in the tasks that we study. 

By replicating the content of one image in an output image that is visually similar to another, our method performs an implicit disentanglement of content and style. In this sense, it is related to recent guided-translation methods, in which the style attributes in the target image are taken from a single image. These include MUNIT~\cite{munit}, the EG-UNIT architecture by~\cite{ma2018exemplar}, and DRIT~\cite{Lee_2018_ECCV}. These methods all employ a style encoding pathway, as well as a content embedding pathway and require a large training set in the target domain. Our method does not employ multiple pathways and does not require such a training set.

\section{Method}
\label{sec:method}
Our method employs an autoencoder for each of the two domains, which are jointly trained in two phases to learn a mapping between the domains. We denote the encoders by $E_A$ and $E_B$ for domains $A$ and $B$ correspondingly, and similarly the decoders by $D_A$ and $D_B$. The cross domain translations are defined based on these encoders and decoders, as done in previous work~\cite{unit,ost}. The transformation between domains $A$ and $B$ is given by $F=E_A \circ D_B$. Similarly, the transformation of a sample $s\in B$ to the first domain is given by the function $G$ as $G(s) = D_A(E_B(s))$. 

Our method involves three main techniques for successfully mapping between the one-shot domain $A$ and the multi-shot domain $B$:

\begin{enumerate}
  \item The training regime comprises two phases. In the first phase, an autoencoder is trained solely for domain $B$. In the second phase, the autoencoder for domain $B$ is further trained along with a separate autoencoder for domain $A$, which is initialized as a clone of the pretrained autoencoder of $B$. This setup (i) helps the training of domain $A$, by utilizing the learned representations of domain $B$ as a prior; and (ii) guides the autoencoder of domain $A$ to acquire a latent representation that aligns to the one captured by domain $B$.
  \item We use Selective Backpropagation to ensure that the encoder and decoder of domain $A$ are adapted to the latent space of domain $B$, and not vice versa. This prevents the latent space from overfitting on the one sample in $A$, and instead be determined by samples in $B$, while $E_A$ and $D_A$ are adapting accordingly. Different from~\cite{ost}, our version of Selective Backpropagation does not involve freezing shared weights. Instead, we simply update a subset of the encoders / decoders with the guideline that $A$ needs to adapt according to $B$.
  \item Unlike ~\cite{ost, unit, munit}, we do not divide the encoders or the decoders to a shared part and an unshared part. Instead, we treat them as four independent networks, which are aligned using a complete set of cycle losses. Most importantly, a feature cycle loss which aligns domain $A$ to domain $B$. This approach enjoys the same benefit as weight sharing, i.e. acquiring a similar latent space for both domains, without the shortcoming of reduced flexibility for the encoder and decoder.   
\end{enumerate}

Due to the limited nature of the data, it is beneficial to employ a data augmentation mechanism. In order to provide a direct comparison, we use identical augmentations to that of~\cite{ost}, which consists of small random rotations of the image and a horizontal translation. Denote by $P(S)$ the augmented training set that is obtained by randomly perturbing the samples of the training set $S \subset B$. In the same manner, we denote $P(x)$ the augmented training set obtained from the single training sample $x$. 

\subsection{Phase I of training}

In the first training phase, we train an autoencoder for domain $B$. The most basic requirement from the autoencoder is reconstruction, i.e. for all $s \in B$, $s \approx D_B(E_B(s))$. In addition, we also require the latent space of $B$ to approximately distribute Gaussian, by using the variational loss. This requirement will later help domain $A$ to align well to domain $B$.    

The total loss for phase I of training is therefore: $\mathcal{L}^{B} = \mathcal{L}_{REC_B} + \lambda_1 \mathcal{L}_{VAE_B}$, where $\lambda_1$ is a weight parameter and
\begin{align}
\mathcal{L}_{REC_B} =&  \sum_{s\in P(S)} \|D_B(E_B(s))- s\|_1  \\ 
\mathcal{L}_{VAE_B} =& \sum_{s\in P(S)} \text{KL}( \{E_B(s) | s\in P(S)\} || \mathcal{N}(0,I) )
\end{align}
where the first loss is the reconstruction loss and the second is the variational loss.

\begin{table*}[t]
\centering

\begin{tabular}{M{1.6cm}M{0.9cm}M{0.9cm}M{1.17cm}M{1.17cm}M{1.6cm}M{1.6cm}M{0.9cm}M{0.9cm}M{0.9cm}M{1.17cm}}
\toprule

	Method & Cycle A & Cycle B & Identity & Variational & Feature Cycle A & Feature Cycle B & GAN A & GAN B & GAN Z & Weight Sharing \\ 
	
	\midrule
	\small{CycleGAN}~\cite{CycleGAN2017} & \checkmark &  \checkmark & \checkmark  & & & & \checkmark & \checkmark & &  \\
	
    \small{MUNIT}~\cite{munit}  & \checkmark & \checkmark & \checkmark  & \checkmark & disentangled & disentangled & \checkmark & \checkmark & & \checkmark  \\
    
    \small{OST}~\cite{ost}  & \checkmark & & \checkmark & \checkmark& & & & \checkmark & & \checkmark \\
    
    \small{I2I}~\cite{i2i_cvpr_2018}  & \checkmark & \checkmark  & \checkmark  & & & & \checkmark  &\checkmark  &\checkmark &\checkmark  \\
     
     \midrule Ours & \checkmark & \checkmark &\checkmark & \checkmark &  & unified & & &   \\
	 \bottomrule
\end{tabular}

\caption{Comparison of losses and network architecture between the proposed method and our baselines. Feature cycle A (resp. B) is the feature cycle loss for images encoded from domain $A$. GAN Z is the loss used by~\cite{i2i_cvpr_2018} for domain confusion in the latent space. GAN A (resp. B) is the GAN loss requiring the translated images from domain $B$ to match the distribution of domain $A$. "Disentangled" refers to the feature cycle presented in~\cite{munit}, where there are separated encoders for style and content, which results in two different latent spaces correspondingly. Our feature cycle involves a unified latent space.  }
\label{tab:compare_baselines}
\end{table*}

\subsection{Phase II of training}

In phase II, we train the autoencoder of domain $A$ jointly with the pretrained autoencoder of domain $B$. We initialize $E_A$ and $D_A$ from $E_B$ and $D_B$ respectively. This initialization guides the autoencoder of $A$ to obtain a representation for $P(x)$ which aligns well to the latent space of $B$. In addition, assuming that the two domains share a similar overall structure, the learned prior of domain $B$ helps in training a robust autoencoder for the one-shot domain.

During this phase, we minimize the following loss:
\begin{align}
\mathcal{L}^{AB} = & \mathcal{L}_{REC_B} +\lambda_2\mathcal{L}_{REC_A} + \lambda_3 \mathcal{L}_{VAE_B} + \lambda_4 \mathcal{L}_{VAE_A} \nonumber \\
&  + \lambda_5 \mathcal{L}_{\text{bab-cycle}} + \lambda_6 \mathcal{L}_{\text{aba-cycle}} + \lambda_7 \mathcal{L}_{\text{f-cycle}}
\end{align}
where $\lambda_i$ are trade-off parameters and the additional loss terms are defined as:

\begin{align}
\mathcal{L}_{REC_A} =&  \sum_{t \in P(x)} \|D_A(E_A(t))- t\|_1\label{eq:recon_a_v2}  
\end{align}
\begin{align}
\mathcal{L}_{VAE_A} =& \sum_{t\in P(x)} \text{KL}( \{E_A(t) | t\in P(x)\} || \mathcal{N}(0,I) ) \label{eq:kl_a_v2}\\
\mathcal{L}_{\text{bab-cycle}} =& \sum_{s\in P(S)} \|D_B(\overline{E_A}(\overline{D_A}(E_B(s))))- s\|_1\label{eq:cyclebab_v2}\\  
\mathcal{L}_{\text{aba-cycle}} =& \sum_{t\in P(x)} \|D_A(\overline{E_B}(\overline{D_B}(E_A(t))))- t\|_1\label{eq:cycleaba_v2}\\  
\mathcal{L}_{\text{f-cycle}} =& \sum_{s\in P(S)} \|E_A(D_A(\overline{E_B}(s)))- \overline{E_B}(s)\|_1\label{eq:shortcycle_v2}
\end{align}

In the above terms, the bar is used to indicate that this network is not updated during backpropogation (``detached'') of this loss. In this way, for example, overfitting to the one sample $x$ is prevented in the autoencoder of $B$.

Losses (\ref{eq:recon_a_v2}) and (\ref{eq:kl_a_v2}) are the analogous losses to those used to pre-train the autoencoder of $B$. Losses (\ref{eq:cyclebab_v2}) and (\ref{eq:cycleaba_v2}) are cycle-consistency losses: from domain $B$ to domain $A$ and back, and from $A$ to $B$ and back. Loss (\ref{eq:shortcycle_v2}) is the short (feature) cycle from the encoded version of samples $s \in B$ to samples in $A$ and back to the latent space. Fig. \ref{fig:arch} depicts the three cycle losses used in our method to obtain an alignment between the two asymmetrical domains.

\subsection{A discussion of the loss terms}
\label{sec:fcycle}

Tab.~\ref{tab:compare_baselines} summarizes the differences in losses and network architecture between the proposed method and the baselines, as well as a recent domain adaptation work (a different but related task) called I2I work~\cite{i2i_cvpr_2018}. 

Over time, the literature shows a tendency to add more losses. However, many of these are not relevant or are detrimental to the one-shot case and our method is considerably simpler. Adversarial losses for domain $A$ are not applicable, because we cannot approximate the distribution of a one-shot domain. We also found in that case, GAN loss for the multi-shot domain does not benefit the transformation. 

The feature-cycle loss ensures the conservation of B features after going through A's decoder and encoder back to the latent space. This improves the alignment between domains and compensates for the unshared weights. Note that unlike ~\cite{munit}, (i) the feature cycle loss is not disentangled between style and content; and (ii) we do not apply this loss for features coming from $x \in A$, since we want the encoder and decoder of domain $A$ to adapt to domain $B$'s latent space, and not vice versa. Moreover, during training, we "freeze" the weights of $E_B$ and backprop only through $D_A$ and $E_A$.  

To achieve a good bidirectional mapping, we want $x \in A$ to be mapped into the same manifold of every $s \in B$. Otherwise, images encoded from domain $A$ will be decoded to domain $B$ in an unmeaningful way, and vice versa. The variational loss in phase I of training forces the latent space of domain $B$ to distribute approximately Gaussian, and in phase II samples from domain $A$ are also adapted to this Gaussian manifold.

\subsection{Network architecture and implementation}
We consider $x \in A$ and samples in B to be images in $\mathbb{R}^{3\times 256\times 256}$. We adopt the successful architecture of Johnson \etal~\cite{Johnson2016Perceptual} for the encoders and decoders. The encoders consist of two 2-stride convolutions and one residual block for digits experiments or four blocks for other experiments, after the convolutional layers. The decoders similarly consist of one or four residual blocks before two deconvolutional layers. Batch normalization and ReLU activations are used between layers. Differing from our baselines~\cite{ost,CycleGAN2017,munit}, we do not use adversarial training and do not employ any discriminator.

For the trade-off parameters associated with the loss terms, we use $\lambda_2$ = $\lambda_5$ = $\lambda_6$ = 1, and $\lambda_3$ = $\lambda_4$ = $\lambda_7$ = 0.001.

\section{Experiments}

\begin{table}[t]
\centering
\begin{tabular}{c@{~}l@{~~}l@{~~}l@{~~}l@{~~~}l@{~~}l@{~~}l}
\toprule
& & \multicolumn{4}{@{}c@{}}{One-Shot} & \multicolumn{2}{@{}c@{}}{All Samples}\\
\cmidrule(l{2pt}r{7pt}){3-6}
 \cmidrule(l{2pt}r{2pt}){7-8}
 
	$A$ & Map & Ours & OST & Cycle & MUNIT & Cycle & MUNIT \\ 
	\midrule
	{\small MNIST} & A$\veryshortarrow$B & 66.50 & 23.50 & 12.00 & 60.50 & 21.46 & 70.81 \\ 
	{\small MNIST} & B$\veryshortarrow$A & 30.73 & 20.82 & 12.34 & 25.22 & 19.32 & 23.58 \\ 
	{\small SVHN} & A$\veryshortarrow$B & 30.00 & 23.50 & 10.50 & 22.00 & 16.54 & 23.25 \\ 
	{\small SVHN} & B$\veryshortarrow$A & 69.48 & 26.58 & 10.80 & 48.06 & 23.60 & 69.11 \\ 
	 \bottomrule
\end{tabular}

\caption{Accuracy for translation from MNIST to SVHN and in the other direction. $A$ is the domain with the one sample. cycle=CycleGAN}
\label{tab:mnis}
\end{table}

\begin{table}

\begin{tabular}{@{~}lllll}
\toprule
 & \multicolumn{2}{c}{Blond $\rightarrow$ Black}&\multicolumn{2}{c}{Black $\rightarrow$ Blond}\\
 \cmidrule(l{7pt}r{7pt}){2-3}
 \cmidrule(l{7pt}r{7pt}){4-5}
 
 Method    &  Cosine & Separation &  Cosine & Separation \\
  &  Sim. & Accuracy &  Sim. & Accuracy\\
\midrule  
OST & 0.39	&0.65	&0.41	&0.67\\
Ours & 0.53	&0.92	&0.54	&0.97\\
\bottomrule
\end{tabular}

\caption{CelebA mapping results using the VGG face descriptor.}
\label{tab:vggfaces}
\end{table}

\begin{table*}
\begin{minipage}[c]{0.497\linewidth}

\centering
\begin{tabular}{@{}llcccc@{}}
\toprule
	 & Method & Summer2 & Winter2 & Monet2 & Photo2 \\ 
	 &  & Winter & Summer & Photo & Monet
\\
\midrule
\parbox[t]{2mm}{\multirow{4}{*}{\rotatebox[origin=c]{90}{Content}}} & OST & 10.25 & 6.84 & 8.62 & 2.09 \\ 
	 & MUNIT & 9.20 & 9.10 & 7.27 & 8.06 \\ 
	 & CycleGAN & 3.07 & 3.74 & 2.56 & 2.35 \\ 
	 & Ours & 1.33 & 1.21 & 2.06 & 1.91 \\ 
	 \midrule
\parbox[t]{2mm}{\multirow{4}{*}{\rotatebox[origin=c]{90}{Style}}} & OST & 8.20 & 2.27 & 6.54 & 3.53 \\ 
	 & MUNIT & 4.10 & 2.83 & 3.44 & 2.65 \\ 
	 & CycleGAN & 3.20 & 2.51 & 1.96 & 2.52 \\ 
	 & Ours & 1.78 & 3.21 & 2.93 & 1.74 \\ 
\bottomrule	 
\end{tabular}
\caption{Mapping from the one-shot domain A to domain B. Both content and style differences are shown for multiple methods. While CycleGAN achieves a lower style difference in some cases, this is obtained for mostly unrelated content, see Fig.~\ref{fig:s2w}(e) and Fig.~\ref{fig:s2w}(j).}
\label{tab:styleatob}

\end{minipage}
~~~\begin{minipage}[c]{0.497\linewidth}

\centering
\begin{tabular}{@{}llcccc@{}}
\toprule
& Method & Summer2 & Winter2 & Monet2 & Photo2 \\ 
	 &  & Winter & Summer & Photo & Monet\\
\midrule
\parbox[t]{2mm}{\multirow{4}{*}{\rotatebox[origin=c]{90}{Content}}}	& OST & 7.32 & 6.02 & 5.71 & 6.48 \\ 
	 & MUNIT & 8.69 & 9.07 & 8.34 & 7.44 \\ 
	 & CycleGAN & 7.53 & 7.90 & 5.83 & 6.73 \\ 
	 & Ours & 1.91 & 1.86 & 3.68 & 3.91 \\ 
	 \midrule
\parbox[t]{2mm}{\multirow{4}{*}{\rotatebox[origin=c]{90}{Style}}}& OST & 4.79 & 6.19 & 9.74 & 9.26 \\ 
	 & MUNIT & 99.10 & 16.58 & 4.62 & 26.24 \\ 
	 & CycleGAN & 6.72 & 13.22 & 8.70 & 10.99 \\ 
	 & Ours & 4.20 & 9.12 & 4.55 & 7.13 \\ 
	 \bottomrule
\end{tabular}
\caption{The results of mapping from domain B to the one-shot domain A. Our method is the only one to achieve both a low content difference and a low style difference. In this more challenging direction, in all cases, our method outperforms the baseline methods.}
\label{tab:stylebtoa}

\end{minipage}
\end{table*}

\begin{table*}
  \centering
\begin{tabular}{l@{~}l@{~}lcccc|l@{~}l@{~}lcccc}
\toprule
  & & & \multicolumn{2}{c}{A is MNIST} & \multicolumn{2}{c|}{A is SVHN}& & & & \multicolumn{2}{c}{A is MNIST} & \multicolumn{2}{c}{A is SVHN}\\
  $\mathcal{L}_\text{f-bab}$ & $\mathcal{L}_\text{f-aba}$ & $\mathcal{L}_\text{f-cycle}$  & $A\veryshortarrow B$ & $B\veryshortarrow A$ & $A\veryshortarrow B$ & $B\veryshortarrow A$ & $\mathcal{L}_\text{f-bab}$ & $\mathcal{L}_\text{f-aba}$ & $\mathcal{L}_\text{f-cycle}$  & $A\veryshortarrow B$ & $B\veryshortarrow A$ & $A\veryshortarrow B$ & $B\veryshortarrow A$\\
 \midrule                          
     0   &  0  &   0   &   39.00  & 25.40 & 25.50   & 30.61  &    1   &  1  &   0   &  43.50  & 26.20 & 18.00 & 48.25 \\
     1   &  0  &   0   &   55.50  & 29.85 & 25.00  & 65.75  &    1   &  0  &   1   &  63.00  & 30.02 & 28.50 & 68.12\\
     0   &  1  &   0   &   36.50  & 24.65 & 22.00  & 26.50  &    0   &  1  &   1   &  63.50  & 26.15 & 30.00 & 17.80 \\
     0   &  0  &   1   &   61.50  & 25.40 & 29.50  & 17.15  &    1   &  1  &   1   &  66.50  & 30.73& 30.00 & 69.48 \\
\bottomrule 
\end{tabular}
  \caption{Ablation study for one-shot MNIST to SVHN translation (all four possibilities), in which we turn on and off the circularity terms.}
    \label{tab:ablation}

\end{table*}

We compare our method to multiple baseline methods from the literature, including OST~\cite{ost}, MUNIT~\cite{munit}, CycleGAN~\cite{CycleGAN2017} and the style transfer method by~\cite{styletransfer}. We provide both qualitative and quantitative results. For the latter, we use the accuracy of classifying the output images as a reliable measure and objective that is often used in previous work. In addition, following the literature, we define style and content losses. Style and content form a tradeoff, by simply copying the image from the target domain, one can obtain perfect style. However, the results indicate clearly that our method is the only one to provide results that have both a small content distance from the source image and a style distance from the target image. Lastly, we provide an ablation analysis, for studying the relative importance of the various components of our method.

Since there is only one sample $x$ from domain $A$, the experiments need to rerun multiple times. We run each experiment 200 times on the digit experiments (sampling a new $x$ each time) and 50 on the other datasets and report the average number obtained.  

\noindent{\bf MNIST to SVHN translation}
Since our method can map in both directions, there are four directions in which the experiments are conducted. For example, we can translate the one-shot MNIST~\cite{mnist} image of a digit to a Street View House Number (SVHN)~\cite{svhn} image. Using the same trained model, we can translate all the SVHN images to the domain defined by the sample $x$ from MNIST. In addition, we can repeat these experiments in the other direction: taking SVHN to be the one-shot domain, and MNIST to be domain $B$.

When transforming an image from MNIST to SVHN, we use a pretrained classifier for SVHN, to predict a label for the translated image and compare it to the label of the source image in MNIST. Note that the MNIST classifier is more limited than the SVHN classifier, since it observed a homogeneous training set. As a result, accuracy is typically higher when translating to SVHN (where the SVHN classifier provides the accuracy) than in the other way. 

Tab.~\ref{tab:mnis} presents the results of translating from and to the one shot domain, for either choice of the one-shot domain. 
As can be seen, the obtained one-shot translation outperforms all baseline methods by a large margin. \textcolor{black} {In addition, it also matches the performance of the baseline methods, when these employ all of the training samples in $A$. Note that both MUNIT and CycleGAN obtain good visual quality in the all sample cases. However, MUNIT is more accurate preserving content, since it disentangles style and content.}

\begin{figure*}
  \centering
  \begin{tabular}{l@{~}c@{~}c@{~}c@{~}c}
\rotatebox[origin=l]{90}{OST} & \includegraphics[width=0.2349\linewidth, clip]{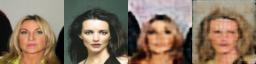} & \includegraphics[width=0.2349\linewidth, clip]{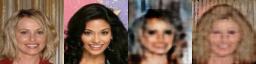} &  \includegraphics[width=0.2349\linewidth, clip]{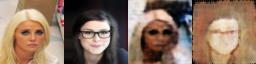}&  \includegraphics[width=0.2349\linewidth, clip]{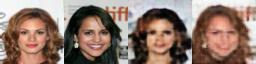}  \\

\rotatebox[origin=l]{90}{Ours} & \includegraphics[width=0.2349\linewidth, clip]{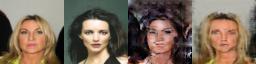} & \includegraphics[width=0.2349\linewidth, clip]{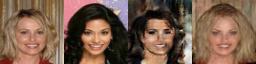} &  \includegraphics[width=0.2349\linewidth, clip]{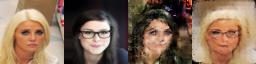} &  \includegraphics[width=0.2349\linewidth, clip]{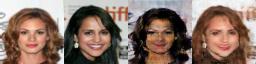}\\
\end{tabular}

\medskip

\begin{tabular}{l@{~}c@{~}c@{~}c@{~}c}
\rotatebox[origin=l]{90}{OST} & \includegraphics[width=0.2349\linewidth, clip]{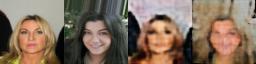} & \includegraphics[width=0.2349\linewidth, clip]{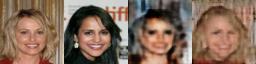} &  \includegraphics[width=0.2349\linewidth, clip]{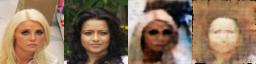}&  \includegraphics[width=0.2349\linewidth, clip]{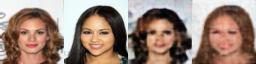}  \\

\rotatebox[origin=l]{90}{Ours} & \includegraphics[width=0.2349\linewidth, clip]{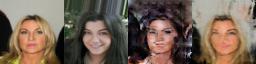} & \includegraphics[width=0.2349\linewidth, clip]{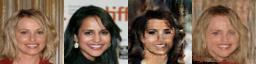} &  \includegraphics[width=0.2349\linewidth, clip]{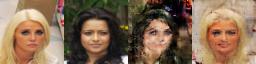} &  \includegraphics[width=0.2349\linewidth, clip]{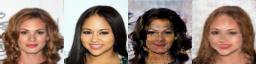}\\
\end{tabular}

  \caption{Each group of images shows the one image $x$ from domain A, a sample image $s$ from domain B, the translation of $x$ to domain B and the translation of $s$ to domain A. (top) OST; (bottom) our method.}
  \vspace{-1.5cm}
  \label{fig:celeba}
\end{figure*}

\begin{figure*}

{\includegraphics[width=\linewidth]{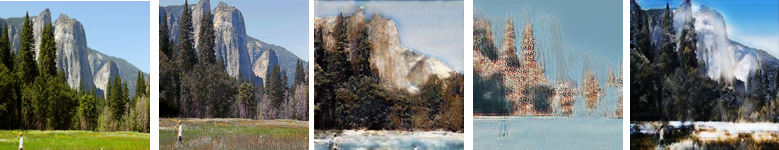}} \\
    \begin{tabular}{@{}c@{}c@{}c@{}c@{}c@{}}
     \kern .1\linewidth(a)&\kern .16782\linewidth(b)&\kern .16782\linewidth(c)&\kern .176782\linewidth(d)&\kern .176782\linewidth(e)
     \end{tabular}\\
     {\includegraphics[width=\linewidth]{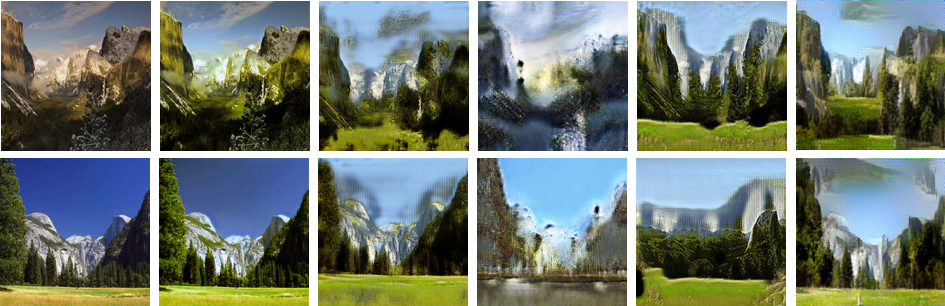}} \\
         \begin{tabular}{@{}c@{}c@{}c@{}c@{}c@{}c@{}}
\kern .081\linewidth(f)&\kern .1326782\linewidth(g)&\kern .14326782\linewidth(h)&\kern .143276782\linewidth(i)&\kern .14376782\linewidth(j)&\kern .14376782\linewidth(k)\end{tabular}
\caption{Mapping summer (domain A) to winter (domain B). (a) The sample $x\in A$. (b) Our result for mapping to $B$. (c) The result of OST. (d) The result of MUNIT. (e) The result of CycleGAN. (f) Two samples $s\in B$. (g) Our result mapping in the other direction, using the same learned model. (h-j) This mapping for the baseline methods in the same order as above. (k) The results of~\cite{styletransfer}, which are only shown in this direction since they cannot benefit from multiple images in $B$.} 

\label{fig:s2w}
\end{figure*}

\begin{figure*}
  \centering
  \begin{tabular}{@{}c@{~}c@{~}c@{~}c@{~}c@{~}c@{}}
\includegraphics[width=0.162249\linewidth, clip]
{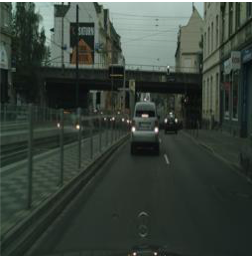} &\includegraphics[width=0.162249\linewidth, clip]
{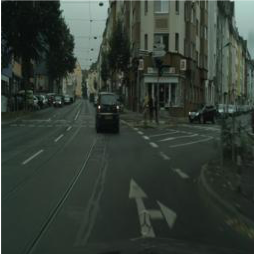} &
\includegraphics[width=0.162249\linewidth, clip]
{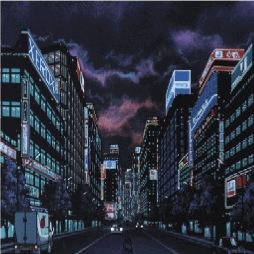} &\includegraphics[width=0.162249\linewidth, clip]
{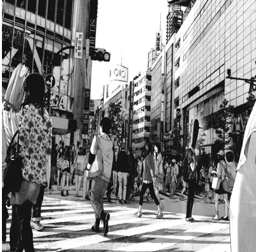} &
\includegraphics[width=0.162249\linewidth, clip]
{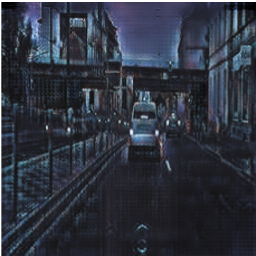} &\includegraphics[width=0.162249\linewidth, clip]
{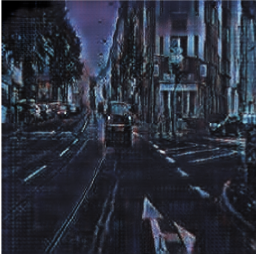} \\
(a) & (b) & (c) & (d) & (e) & (f)\\
\includegraphics[width=0.162249\linewidth, clip]
{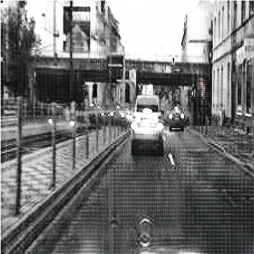} &\includegraphics[width=0.162249\linewidth, clip]
{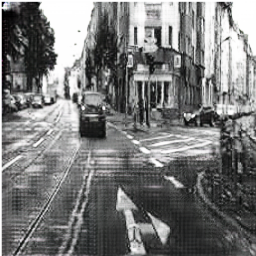} &
\includegraphics[width=0.162249\linewidth, clip]
{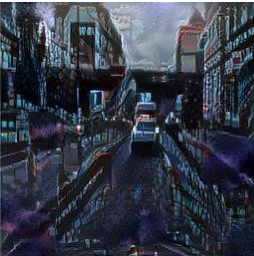} &\includegraphics[width=0.162249\linewidth, clip]
{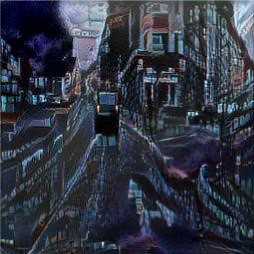} &
\includegraphics[width=0.162249\linewidth, clip]
{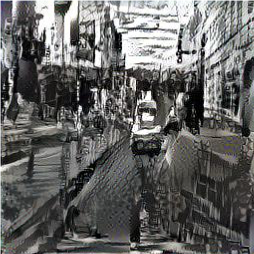} &\includegraphics[width=0.162249\linewidth, clip]
{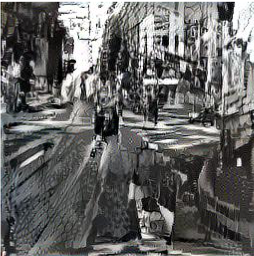} \\
(g) & (h) & (i) & (j) & (k) & (l) \\
\end{tabular}
  \caption{Style transfer results of street scene images. (a,b) content images. (c,d) two style images. (e-h) our results obtained by combining the content and the style images. (i-l) the same results for~\cite{styletransfer}. }
  \label{fig:style}
\end{figure*}

\noindent{\bf CelebA experiments } 
The CelebA dataset~\cite{celeba, celeba2} was annotated for multiple attributes, including three properties that are often used for testing domain translation methods: the person's gender, hair color, and the existence of glasses in the image. Out of these three, male to female does not make sense in the one-shot scenario, since the distributions overlap, and no method was able to add or remove glasses, after seeing one sample of a person with glasses. This is because these are interpreted by the networks, as part of the skin around the eyes (e.g., dark circles under the eyes). 

Fig.~\ref{fig:celeba} presents typical results obtained by our method and OST for translating between blond and black hair. It is evident that the baseline method does not present as convincing a translation as our method.

In order to quantify the quality of the face translation that was performed, we use the representation layer of VGG faces~\cite{Parkhi15} on the image in A and its output in B. The content would be transformed well, if the VGG representation of the face remains unchanged, since it is trained to be invariant to hair color. As is typically done, we employ the cosine similarities to compare two face representation vectors. The results are shown in Tab.~\ref{tab:vggfaces}. As can be seen, our method obtains a higher similarity than the baseline method.

The face descriptor metric captures the content that is being transferred. Our ability to create images that are faithful to the target domain is evaluated by employing a classifier that is trained to distinguish between images of blonds and those of people with black hair. Tab.~\ref{tab:vggfaces} reports the ratio of cases for which the classifier assigned the translated image to the target class. As can be seen, the classifier assigns our output image to the target class much more  frequently than it does for the baseline method.

\noindent{\bf Photo translation tasks}
We next consider the task of two-way translation from real images to the paintings of Monet~\cite{CycleGAN2017} and between summer and winter images, following~\cite{CycleGAN2017}. To assess the quality of these translations, we measure the perceptual distance~\cite{Johnson2016Perceptual} between the source image and the translated version. A low value is taken in the literature as an indication that much of the content is preserved. To compare the style differences between the translated images and target domain images, we employ the Gram metric, as used by style transfer methods~\cite{styletransfer}. 

Tab.~\ref{tab:styleatob} reports the obtained score, when mapping from $A$, the one-shot domain, to $B$. Tab~\ref{tab:stylebtoa} reports the scores obtained, when mapping in the other direction. As can be seen, our method presents preferable scores over all baselines, with the one exception of CycleGAN obtaining a lower style loss in Monet2Photo. However, CycleGAN completely fails in replicating the image's content. Sample results obtained with each method are shown in Fig.~\ref{fig:s2w} for summer to winter (see supplementary for the other domains).The figure also compares visually with~\cite{styletransfer}. As can be seen, our results present a translation that is more faithful to the content of the source image and less distorted than the baseline methods.

Finally, in Fig.~\ref{fig:style}, we test our method for the task of style transfer on more extreme out-of-dataset style images (taken as the one-shot domain) and compare with~\cite{styletransfer}. Note that the baseline method does not benefit from having a training set from the source domain. However, it employs a classifier that was trained on imagenet. As can be seen, the translation performed by our method preserves the image content, while being able to transfer the style. The baseline method distorts the content, c.f., the straight lines in the content image.

\subsection{Ablation analysis}

{\color{black}{
\noindent{\bf Relative importance of cycle losses}
Since we introduce the $\mathcal{L}_\text{f-cycle}$, and the baseline method of OST employs only one cycle, we focus the ablation analysis on understanding the relative importance of each cycle loss of the three. The results of removing some of the cycles are presented in Tab.~\ref{tab:ablation}. As can be seen, the feature cycle we introduce makes a great contribution to the mapping from $B$ to $A$ and removing it is detrimental.
}}

\begin{figure*}
  \centering
  \begin{tabular}{c@{~}c@{~}c@{~}c@{~}c}
\raisebox{0.4cm}{(a)}
 & \includegraphics[width=0.23\linewidth, clip]{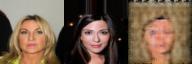} & \includegraphics[width=0.23\linewidth, clip]{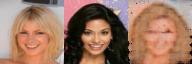} &  \includegraphics[width=0.23\linewidth, clip]{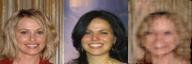} &  \includegraphics[width=0.23\linewidth, clip]{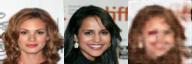} \\
\raisebox{0.4cm}{(b)}
  & \includegraphics[width=0.23\linewidth, clip]{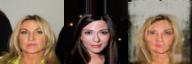} & \includegraphics[width=0.23\linewidth, clip]{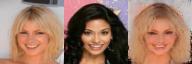} &  \includegraphics[width=0.23\linewidth, clip]{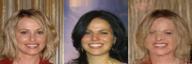} & \includegraphics[width=0.23\linewidth, clip]{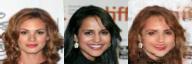}   \\
\end{tabular}
  \caption{Each group of images shows the one image $x$ from domain A, a sample image $s$ from domain B, and the translation of $s$ to domain A. (a) our method with weight sharing; (b) our method with independent weights per domain. Using independent weights per domain allows better flexability and results in more accurate and less blurry mappings.}
  \label{fig:weight_sharing}
\end{figure*}

\begin{figure}
  \centering
 \begin{tabular}{@{}c@{~}c@{~}c@{}}
\includegraphics[height=0.0871125\linewidth, clip]{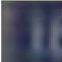} & \includegraphics[height=0.0871125\linewidth, clip]{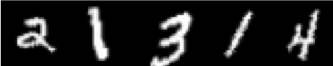} &\includegraphics[height=0.0871125\linewidth, clip]{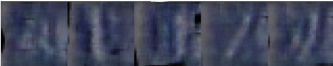} \\
(a) & (b) & (c)\\
& \includegraphics[height=0.0871125\linewidth, clip]{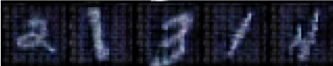} &
\includegraphics[height=0.0871125\linewidth, clip]{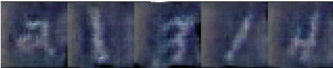}  \\
& (d) & (e)\\
\end{tabular}

  \caption{Mapping from domain $B$ to domain $A$. (a) one-shot image from SVHN (domain $A$). (b) sample images from MNIST (domain $B$). (c) mapping $B$ to $A$ without feature cycle loss. (d) mapping with random feature cycle loss. (e) mapping with feature cycle loss. The feature cycle loss ensures the conservation of features for domain $B$, when transforming to domain $A$. A random feature cycle also ensures this, but to a lesser degree.}
  \label{fig:rand_f_cycle}
\end{figure}

{\color{black}{
\noindent{\bf Further analysis of the feature cycle}
The feature cycle can be interpreted as a reconstruction loss of virtual samples from domain $A$ in feature space. As explained in Sec.~\ref{sec:method}, in phase II, we freeze $E_B(s)$ and, therefore, this encoding acts as a "ground truth" feature representation of a sample from domain $B$. Then, we obtain $D_A(E_B(s))$, which can be viewed as a pseudo $A$ sample, generated by mapping the $B$ sample into domain $A$. Finally, we bring this sample back to the latent space and require $E_A(D_A(E_B(s))) \approx E_B(s)$, which is the equivalent of demanding the reconstruction of the generated $A$ sample, but in the feature space. This shorter cycle is more efficient than passing through $D_A$ once again for a reconstruction at pixel space, since we already have the "ground truth" features $E_B(s)$, unlike the usual case of pixel reconstruction loss. 

This analysis has three implications: (i) generation of multiple ($|S|$) new samples for the one-shot domain $A$ and a reconstruction loss for these samples; (ii) the latent space of $A$ for these reconstructions is the same exact latent space as samples from $B$; and (iii) conservation of domain $B$ features, when passing through the encoder/decoder of domain $A$. 

To verify our claim, we tested a cycle of random permutations of latent codes from $B$. Fig.~\ref{fig:rand_f_cycle} presents an example for the mapping of samples from domain $B$ (MNIST) to domain $A$ (SVHN) with $\mathcal{L}_\text{f-cycle}$, without $\mathcal{L}_\text{f-cycle}$, and with a feature cycle of random permutations of $E_B(s)$. As can be seen, the random feature cycle "highlights" the features when mapping $B$ to $A$ in an exaggerated way, but nevertheless, improves feature conservation between domains. The normal feature cycle balances the trade-off between feature conservation, and obtaining a mapping which remains true to the target domain style. Numerically, the results of employing random vectors (instead of encoding of images from $B$ for the f-cycle) are mid-way of using and not using f-cycle (The four accuracies reported in Tab.~\ref{tab:ablation} are 65.00,23.53,23.00,63.45 for this case, respectively).

Although the feature cycle loss is highly successful for the one-shot scenario, it is not necessarily suitable for the multi-shot case, where there are many samples from $A$ and we do not need to generate pseudo samples for domain $A$. Therefore, it is not surprising that it is absent from the literature methods. Note that the feature cycle in ~\cite{munit} has different implications, since it employs a disentangled representation between style and content. It, therefore, also plays a role of creating synthetic examples (mixed style and content) for which no real alternatives exist.
}}

\noindent{\bf Weight sharing}
Multiple domain mapping methods ~\cite{munit, i2i_cvpr_2018} found weight sharing to be beneficial. However, we found it to be detrimental for the one-shot scenario. Using separate weights, the encoder and decoder of domain $A$ are more adaptable to domain $B$, while maintaining a correct representation for domain $A$. Fig.~\ref{fig:weight_sharing} illustrates this kind of behavior for the case of mapping between blond (domain $A$) and black (domain $B$) faces. As can be seen, weight sharing restricts the decoders from adapting well between domains and leads to blurry images.

\section{Conclusion}

The problem of cross domain translation is highly researched and the quality of the results is constantly improving. In addition, the field is gradually adding new capabilities that at first seem surprising. The ability to perform the visual translation task in an unsupervised way, was unforeseen by either the machine learning literature or the cognitive science one. It was also not obvious that this translation can be performed in a one-sided way, since all of the first results relied on circularity. The one-shot case from $A$ to $B$ was unexpected, since the training losses that were applied would easily fit on the one sample from $A$. Finally, as noted in the literature, the opposite translation is even more challenging, due to the inability to properly model domain $A$.

We demonstrate not only that this mapping is possible, but also that the algorithm developed for doing so, is substantially more effective than the literature methods, even in the opposite direction. The ability to use one-shot methods to blend a new visual experience with existing visual knowledge, provides a way to accumulate information over time and to project existing knowledge onto a new sample. 

\section*{Acknowledgements}

This project has received funding from the European Research Council (ERC) under the European Union’s Horizon 2020 research and innovation programme (grant ERC CoG 725974). 

\clearpage

{\small
\bibliographystyle{ieee_fullname}
\bibliography{gans}
}

\end{document}